\newcommand{\CLASSINPUTtoptextmargin}{0.77in}
\newcommand{\CLASSINPUTbottomtextmargin}{1.02in}

\documentclass[conference]{IEEEtran}
\IEEEoverridecommandlockouts

\usepackage{amsmath,amsfonts,amssymb}

\usepackage{graphicx}
\usepackage[
    caption=false,
    font=footnotesize,
    labelfont=rm,
    textfont=rm
]{subfig}

\usepackage{array}
\usepackage{booktabs}
\usepackage{multirow}
\usepackage{makecell}
\usepackage{tabularx}
\usepackage{colortbl}

\usepackage{algorithm}
\usepackage{algorithmic}

\usepackage{cite}
\usepackage{xcolor}
\usepackage{textcomp}
\usepackage{url}

\usepackage{cleveref}
\usepackage{multirow, makecell}
\usepackage{colortbl}  
\usepackage{amsmath}
\usepackage{amssymb}
\usepackage{bbding}

\definecolor{Gray}{gray}{0.94}
\definecolor{liGray}{gray}{0.5}
\definecolor{LightCyan}{rgb}{0.88,1,1}

\newlength\savewidth

\usepackage{cite}
\usepackage{amsmath,amssymb,amsfonts}
\usepackage{algorithmic}
\usepackage{graphicx}
\usepackage{textcomp}
\usepackage{xcolor}
\def\BibTeX{{\rm B\kern-.05em{\sc i\kern-.025em b}\kern-.08em
    T\kern-.1667em\lower.7ex\hbox{E}\kern-.125emX}}

\begin{document}

\title{DiffAnchor: Anchor-Space Diffusion for 
Monocular 3D Hand Pose Estimation
}

\author{
\IEEEauthorblockN{
Changlong Jiang\textsuperscript{1},
Youci Tang\textsuperscript{1},
Haohong Kuang\textsuperscript{1},
Hang Xu\textsuperscript{1},
Jinghong Zheng\textsuperscript{1},
Ran Wang\textsuperscript{2}, \\
Zhenghong Yu\textsuperscript{3},
Chen Feng\textsuperscript{4},
Xiaofeng Yue\textsuperscript{4,5}, and
Yang Xiao\textsuperscript{1 \textdagger}
}

\IEEEauthorblockA{
\textsuperscript{1}
National Key Laboratory of Multispectral Information Intelligent Processing Technology,  School of Artificial \\ Intelligence and Automation, Huazhong University of Science and Technology, Wuhan, 430074, China.
}

\IEEEauthorblockA{
\textsuperscript{2}
School of Journalism and Information Communication, 
Huazhong \\University of Science and Technology,
Wuhan, 430074, China.
}

\IEEEauthorblockA{
\textsuperscript{3}
School of  Robotics, Guangdong Polytechnic of Science and Technology,  Zhuhai, 519090, China.}

\IEEEauthorblockA{
\textsuperscript{4}
Wuhan · HGCyber Data System Co., Ltd., Wuhan, China.}


\IEEEauthorblockA{
\textsuperscript{5}
School of Artificial Intelligence and Automation, Huazhong \\
University of Science and Technology, Wuhan, 430074, China.
}


\IEEEauthorblockA{
E-mail: \{cljiang, tangyouci, haohong\_kuang, hang\_xu, deepzheng,  rex\_wang\}@hust.edu.cn, yuzhenghong@gdit.edu.cn, \\ fengchen@hgtech.com.cn, yuexiaofeng@hgcyber.com.cn, Yang\_Xiao@hust.edu.cn. \textsuperscript{\textdagger}Corresponding author.
}

\thanks{This work was supported by the National Natural Science Foundation of China under Grants 62671264 and 62271221, and the Key Research and Development Project of Hubei Province under Grant 2026BEB001.}

}


\maketitle

\begin{abstract}
Accurate and efficient monocular 3D hand pose estimation is an important perception capability for embodied human interaction and dexterous manipulation. Recent diffusion-based methods model pose uncertainty by iteratively denoising 3D joint coordinates. However, directly diffusing the final joints requires the generative process to simultaneously explore plausible poses and recover precise coordinates, often relying on additional denoising steps or multiple hypotheses and consequently increasing inference cost. In this work, we propose DiffAnchor, a few-step anchor-space diffusion framework that diffuses anchors rather than joints. Our key insight is that stochastic exploration does not need to operate directly on the final output. Instead, diffusion searches for informative intermediate anchors that serve as local 3D regressors, while anchor-to-joint offset prediction and weighted aggregation complete precise joint localization. Specifically, fixed context anchors provide stable global references, whereas noisy dynamic anchors are iteratively refined toward informative joint regions. An importance-guided anchor selection and replenishment mechanism further retains useful anchors, removes ineffective ones, and explores insufficiently covered regions. This design reduces the precision burden of the diffusion process and enables accurate prediction using only a few denoising steps and a single hypothesis. Extensive experiments on HO-3D V2, NYU, ICVL, and InterHand2.6M demonstrate state-of-the-art performance on the first three benchmarks and competitive performance on challenging interacting-hand estimation, while maintaining inference efficiency comparable to deterministic regression baselines.
\end{abstract}



\section{Introduction}

\begin{figure}[t]
\centering
\includegraphics[width=0.99\linewidth]{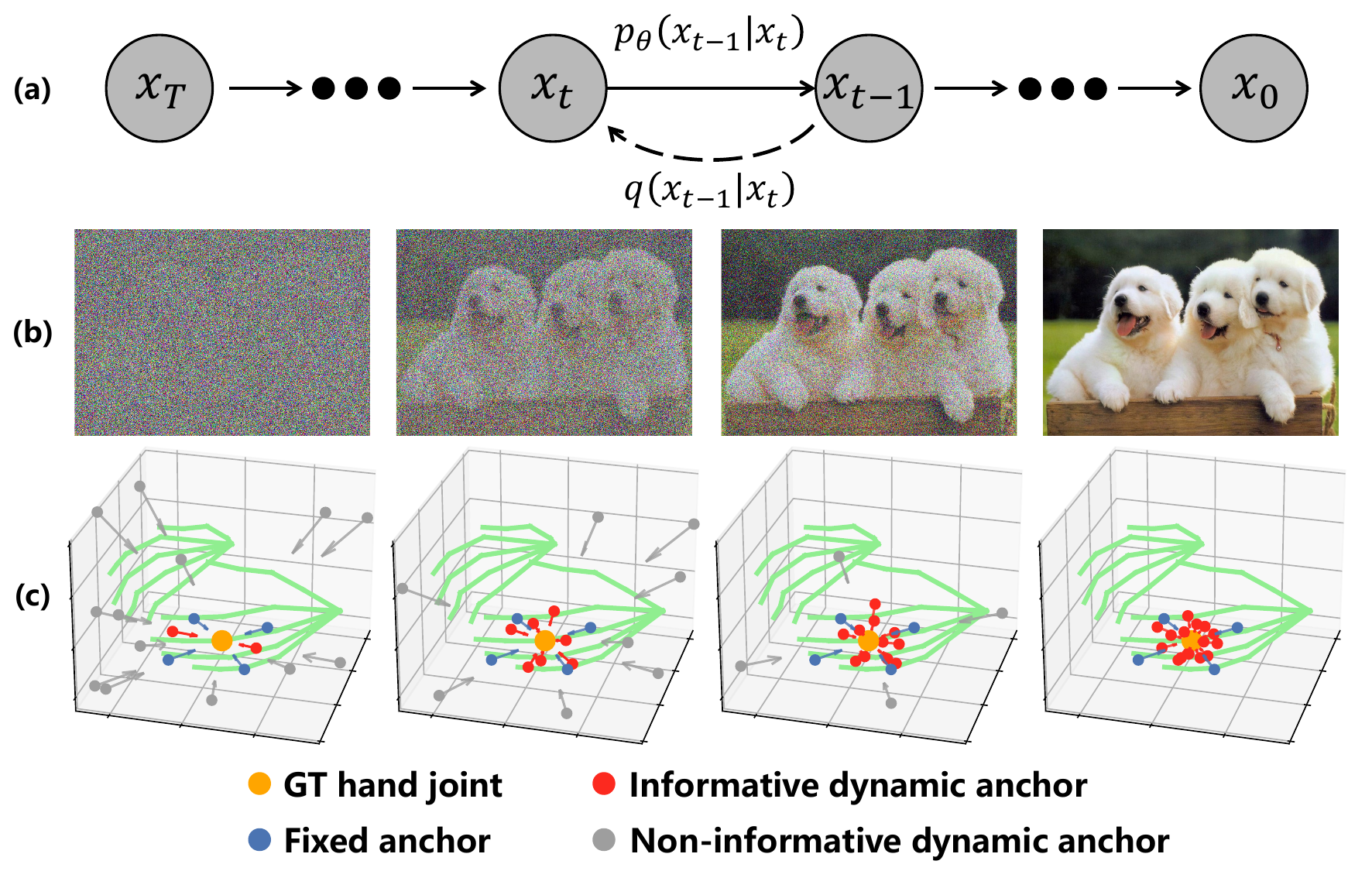}
\vspace{-8pt}
\caption{The main idea of DiffAnchor. (a) A diffusion model with the diffusion process $q$ and the reverse process $p_\theta$. (b) Diffusion model for image generation task. (c) Our anchor-to-joint paradigm with the denoised informative anchors from noisy anchors.} 
\label{fig:main_idea}
\vspace{-8pt}
\end{figure}

\IEEEPARstart{M}{onocular} 3D hand pose estimation aims to recover the 3D locations of hand joints from a single RGB or depth image.
It serves as an important perception capability for human--computer interaction, augmented and virtual reality, and embodied systems involving dexterous manipulation.
Despite substantial progress~\cite{meng20223d,moon2020interhand2,jiang2023a2j,tan2024epmf,ding2023image,huang2020location,niu2022efficient,zhang2020collaborative,liu2021discrimination}, accurate hand pose estimation remains challenging because of severe self-occlusion, hand--hand and hand--object occlusions, locally similar visual patterns, and the inherent depth ambiguity of monocular observations.
These difficulties become particularly pronounced in interacting-hand and hand--object scenarios, where the visual evidence of several joints can be incomplete or even entirely missing.

Diffusion models provide a promising generative formulation for addressing such ambiguity.
Instead of predicting a pose through a single deterministic forward pass, diffusion-based pose estimators progressively recover plausible 3D structures from noisy initializations~\cite{gong2023diffpose,cheng2024handdiff}.
The iterative denoising process allows the model to explore the conditional pose distribution and gradually correct structurally implausible predictions, which is particularly useful when the image admits multiple possible 3D interpretations.

However, existing diffusion-based pose estimators generally define the final joint coordinates as the diffusion variables.
Under this formulation, the same denoising process must accomplish two different objectives: exploring plausible global pose configurations and recovering the precise coordinate of every joint.
The former benefits from stochasticity and a broad search space, whereas the latter requires stable and highly accurate localization.
Since any residual perturbation in the diffused coordinates is directly transferred to the final pose, accurate prediction commonly relies on repeated denoising, and some methods further aggregate multiple pose hypotheses.
These operations introduce additional inference cost, which limits the practical advantage of diffusion for high-precision hand pose estimation.
This raises a key question:
\emph{Must diffusion operate directly on the final joints to exploit its
generative exploration capability?}

As illustrated in \Cref{fig:main_idea}, our answer is no.
We argue that stochastic exploration and precise localization need not be performed in the same prediction space.
Instead of directly diffusing the final joints, we transfer stochastic exploration to an intermediate anchor space.
The generated anchors serve as local 3D regressors: diffusion only needs to move them toward informative joint regions, rather than precisely recover the target coordinates.
Once useful local regions have been identified, anchor-to-joint offset regression and weighted aggregation complete the final joint localization.
In this way, stochastic exploration is performed in the anchor space, whereas precise localization is completed in the joint space.

Based on this insight, we propose \textbf{DiffAnchor}, a few-step anchor-space diffusion model for monocular 3D hand pose estimation.
DiffAnchor contains two complementary types of anchors.
Fixed context anchors provide stable spatial references throughout the inference process, while dynamic anchors are initialized from Gaussian noise and iteratively refined according to the input image.
Each anchor predicts its 3D offsets and importance weights with respect to the hand joints.
The final joint coordinates are then obtained through a weighted anchor-to-joint ensemble, which reduces the sensitivity of the output to the perturbation of any individual dynamic anchor.

Furthermore, not all generated anchors are equally informative.
Some anchors rapidly approach useful joint neighborhoods and provide short-range local regressors, whereas others remain far from the relevant hand regions or produce unreliable offsets.
We therefore introduce an importance-guided anchor selection and replenishment mechanism.
At each denoising step, high-confidence anchors are retained for subsequent refinement, while ineffective anchors are removed and replaced by newly sampled candidates.
The retained anchors exploit already discovered joint regions, and the replenished candidates continue to explore insufficiently covered regions.
This iterative process allows anchors to adapt to different hand poses, occlusion patterns, or interaction conditions.

We evaluate DiffAnchor on four 3D hand pose benchmarks: HO-3D V2~\cite{hampali2020honnotate}, NYU~\cite{tompson2014real}, ICVL~\cite{tang2014latent}, and InterHand2.6M~\cite{moon2020interhand2}.
These datasets cover RGB and depth inputs as well as hand--object interaction, single-hand, and interacting-hand scenarios.
DiffAnchor achieves state-of-the-art performance on HO-3D V2, NYU, and ICVL, and obtains the best model-free two-hand result on InterHand2.6M.
Moreover, it reaches near-saturated accuracy using only a few denoising steps and a single hypothesis, while maintaining inference efficiency comparable to deterministic regression baselines.

Overall, the main contributions are as follows:

$\bullet$ We identify the coupling between stochastic pose exploration and precise coordinate localization in direct joint diffusion.
To address this issue, we propose to diffuse intermediate anchors rather than final joints, separating stochastic exploration in the anchor space from precise localization in the joint space.

$\bullet$ We present DiffAnchor, a few-step anchor-space diffusion framework that combines fixed context anchors, dynamically generated local regressors, and anchor-to-joint offset aggregation for stable and accurate 3D hand pose estimation.

$\bullet$ We introduce an importance-guided anchor selection and replenishment mechanism that retains informative anchors while continually exploring insufficiently covered 3D regions.
Extensive experiments across four benchmarks validate the effectiveness, few-step efficiency, cross-modality applicability, and robustness of the proposed framework.


\section{Related Work}
\label{sec:related_work}

{\bf 3D hand pose estimation.}
Existing methods can be broadly divided into model-based and model-free approaches.
Model-based methods reconstruct hand meshes using parametric hand models~\cite{zhang2021interacting,romero2022embodied,xu2026moeg} or non-parametric mesh representations~\cite{ge20193d,moon2020i2l,li2022interacting,wang2023memahand,li2023renderih}, and subsequently recover the joint coordinates.
Model-free methods directly estimate hand joints through heatmap prediction~\cite{wang2018mask,fan2021learning,hampali2022keypoint,moon2020interhand2} or coordinate regression~\cite{cheng2021handfoldingnet,xiong2019a2j,jiang2023a2j,moon2018v2v,meng20223d,zheng2026pandapose,kuang2026jump}.
Despite substantial progress, monocular hand pose estimation remains challenging under severe self-occlusion, hand--hand or hand--object interactions, and depth ambiguity.
Rather than directly reconstructing a hand mesh or regressing the final joints, DiffAnchor introduces intermediate 3D anchors to explore informative local regions before precise joint localization.

\begin{figure*}[t]
\centering
\includegraphics[width=0.85\linewidth]{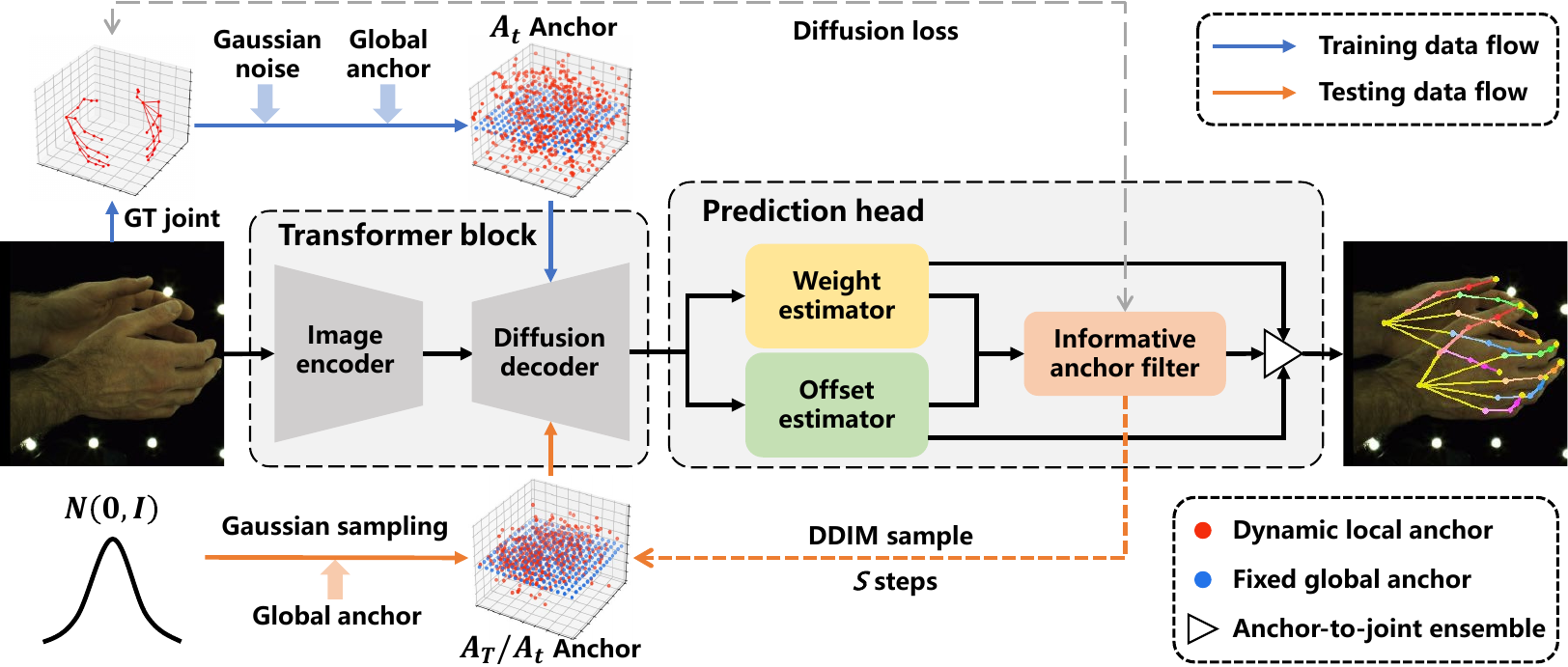}
\vspace{-8pt}
\caption{
Overview of DiffAnchor.
Given an input image, the image encoder extracts multi-scale visual features.
Fixed context anchors provide stable global spatial references, while dynamic anchors are initialized from noise and iteratively refined in the anchor space.
At each denoising step, the diffusion decoder predicts anchor-to-joint offsets and importance scores.
Joint-specific informative anchors are selected for weighted anchor-to-joint aggregation, whereas high-confidence denoised anchors are retained and the remaining anchors are replenished with newly sampled candidates.
After a few iterations, the final 3D hand pose is obtained from the anchor-to-joint ensemble.
}
\label{fig:pipeline}
\end{figure*}

{\bf Anchor-based pose estimation.}
Anchor-based methods decompose direct joint prediction into anchor-to-joint offset regression and weighted proposal aggregation.
A2J~\cite{xiong2019a2j} predicts joint offsets and confidence weights from uniformly preset 3D anchors, while A2J-Transformer~\cite{jiang2023a2j} and A2J-Transformer+~\cite{jiang20253d} further enhance anchor interaction and articulated local regression.
However, existing anchors are generally constructed using preset grids or deterministic rules and cannot adapt their spatial distributions to individual images.
Some anchors may therefore remain far from the target joints, leading to difficult long-range offset regression, whereas increasing anchor density introduces additional computation.
Moreover, ineffective anchors may interfere with weighted aggregation.
In contrast, DiffAnchor treats anchors as generative intermediate variables and iteratively explores, selects, and updates their spatial distribution according to the input image.

{\bf Diffusion-based pose estimation.}
Diffusion models~\cite{ho2020denoising,song2020denoising} have been applied to 3D pose estimation by progressively denoising noisy pose hypotheses~\cite{gong2023diffpose,cheng2024handdiff}.
Their iterative sampling process provides a natural mechanism for modeling ambiguous pose distributions.
However, existing methods generally define the final joint coordinates as diffusion variables.
The same denoising process must therefore both explore plausible global structures and recover highly precise joint coordinates, often requiring additional sampling steps or multiple hypotheses.
DiffAnchor instead transfers stochastic exploration to the intermediate anchor space.
Diffusion only needs to discover informative local regression regions, while anchor-to-joint offset prediction and weighted aggregation complete precise localization, enabling accurate estimation with a few denoising steps and a single hypothesis.

\section{Method}
\label{sec:method}

\subsection{Overview and Problem Formulation}
\label{subsec:overview}

Given a monocular RGB or depth image $\mathbf{I}$, the goal of 3D hand pose estimation is to predict the coordinates of $N_J$ hand joints:
\begin{equation}
    \mathbf{Y}
    =
    [\mathbf{y}_1,\ldots,\mathbf{y}_{N_J}]
    \in
    \mathbb{R}^{N_J\times3},
\end{equation}
where $\mathbf{y}_j\in\mathbb{R}^{3}$ denotes the coordinate of joint $j$.

DiffAnchor performs stochastic exploration in an intermediate anchor space rather than directly diffusing the final joint coordinates.
As illustrated in \Cref{fig:pipeline}, the complete anchor set at denoising step $t$ consists of fixed context anchors and dynamic anchors:
\begin{equation}
    \mathcal{A}_t
    =
    \mathcal{A}^{\mathrm{fix}}
    \cup
    \mathcal{A}_t^{\mathrm{dyn}},
    \qquad
    N_A=N_{\mathrm{fix}}+N_{\mathrm{dyn}},
    \label{eq:anchor_set}
\end{equation}
where
$\mathcal{A}^{\mathrm{fix}}
=
\{\mathbf{a}^{\mathrm{fix}}_i\}_{i=1}^{N_{\mathrm{fix}}}$
remains unchanged throughout inference, while
$\mathcal{A}_t^{\mathrm{dyn}}
=
\{\mathbf{a}_{t,i}^{\mathrm{dyn}}\}_{i=1}^{N_{\mathrm{dyn}}}$
is iteratively refined.

Following the anchor-to-joint formulation~\cite{xiong2019a2j,jiang2023a2j,jiang20253d}, each anchor predicts a 3D offset and an importance logit with respect to every joint.
For anchor $\mathbf{a}_{t,i}$ and joint $j$, the corresponding joint proposal is
\begin{equation}
    \mathbf{p}_{t,i,j}
    =
    \mathbf{a}_{t,i}
    +
    \mathbf{o}_{t,i,j},
    \label{eq:joint_proposal}
\end{equation}
where
$\mathbf{o}_{t,i,j}\in\mathbb{R}^{3}$
is the predicted anchor-to-joint offset.
The final joint coordinates are obtained by aggregating the proposals from informative anchors.

The two anchor types play complementary roles.
Fixed context anchors provide stable global references and prevent the prediction from depending entirely on stochastic initialization.
Dynamic anchors, in contrast, act as image-conditioned local regressors whose spatial distribution is progressively explored and refined.
Consequently, diffusion only needs to discover informative local regression regions, while precise coordinate localization is completed through anchor-to-joint offset prediction and weighted aggregation.

\subsection{Anchor-Space Diffusion}
\label{subsec:anchor_diffusion}

\subsubsection{Training anchor construction}

During training, we construct a dynamic-anchor representation using the ground-truth joints and additional Gaussian anchors.
Specifically,
\begin{equation}
    \mathbf{A}_0^{\mathrm{dyn}}
    =
    \operatorname{Concat}
    \left(
        \mathbf{Y},
        \mathbf{R}
    \right)
    \in
    \mathbb{R}^{N_{\mathrm{dyn}}\times3},
    \label{eq:clean_anchor_set}
\end{equation}
where
\begin{equation}
    \mathbf{R}
    \sim
    \mathcal{N}
    \left(
        \mathbf{0},
        \mathbf{I}
    \right),
    \qquad
    \mathbf{R}
    \in
    \mathbb{R}^{(N_{\mathrm{dyn}}-N_J)\times3}.
\end{equation}
All joint and anchor coordinates are represented in the same normalized 3D coordinate system.

The ground-truth joints provide pose-related spatial references, while the Gaussian anchors expose the model to a broader set of initial locations.
Unlike approaches that require every diffusion variable to reproduce an exact final joint coordinate, DiffAnchor only requires the reverse process to move dynamic anchors toward regions that are useful for local joint regression.

Following the standard forward diffusion process~\cite{ho2020denoising,nichol2021improved}, Gaussian noise is added to the clean dynamic anchors:
\begin{equation}
    q
    \left(
        \mathbf{A}_t^{\mathrm{dyn}}
        \mid
        \mathbf{A}_0^{\mathrm{dyn}}
    \right)
    =
    \mathcal{N}
    \left(
        \sqrt{\bar{\alpha}_t}
        \mathbf{A}_0^{\mathrm{dyn}},
        (1-\bar{\alpha}_t)\mathbf{I}
    \right),
    \label{eq:forward_diffusion}
\end{equation}
or equivalently,
\begin{equation}
    \mathbf{A}_t^{\mathrm{dyn}}
    =
    \sqrt{\bar{\alpha}_t}
    \mathbf{A}_0^{\mathrm{dyn}}
    +
    \sqrt{1-\bar{\alpha}_t}
    \boldsymbol{\epsilon},
    \qquad
    \boldsymbol{\epsilon}
    \sim
    \mathcal{N}(\mathbf{0},\mathbf{I}),
    \label{eq:forward_sampling}
\end{equation}
where
$\bar{\alpha}_t=\prod_{s=1}^{t}(1-\beta_s)$
and $\{\beta_s\}$ denotes the variance schedule.
We adopt the cosine noise schedule in~\cite{nichol2021improved}.
The fixed anchors are not perturbed and remain unchanged for all diffusion timesteps.

\subsubsection{Image-conditioned reverse process}

Given the noisy dynamic anchors
$\mathbf{A}_t^{\mathrm{dyn}}$,
the complete input anchor set is formed as
\begin{equation}
    \mathbf{A}_t
    =
    \operatorname{Concat}
    \left(
        \mathbf{A}^{\mathrm{fix}},
        \mathbf{A}_t^{\mathrm{dyn}}
    \right).
\end{equation}

The reverse process is conditioned on both the input image and the current anchor distribution.
Instead of directly predicting the final joint coordinates as the denoised variables, the network predicts anchor-to-joint offsets and importance logits:
\begin{equation}
    \left(
        \mathbf{O}_t,
        \mathbf{W}_t
    \right)
    =
    f_{\theta}
    \left(
        \mathbf{I},
        \mathbf{A}_t,
        t
    \right),
    \label{eq:network_prediction}
\end{equation}
where
\begin{equation}
    \mathbf{O}_t
    \in
    \mathbb{R}^{N_A\times N_J\times3},
    \qquad
    \mathbf{W}_t
    \in
    \mathbb{R}^{N_A\times N_J}.
\end{equation}

For every dynamic anchor, the predicted offsets are used to construct a denoised anchor estimate.
This estimate serves as the clean-anchor prediction required by the subsequent DDIM update.
The detailed parent-joint assignment and anchor update procedure are introduced in
\Cref{subsec:anchor_update}.

\begin{figure}[t]
\centering
\includegraphics[width=0.9\linewidth]{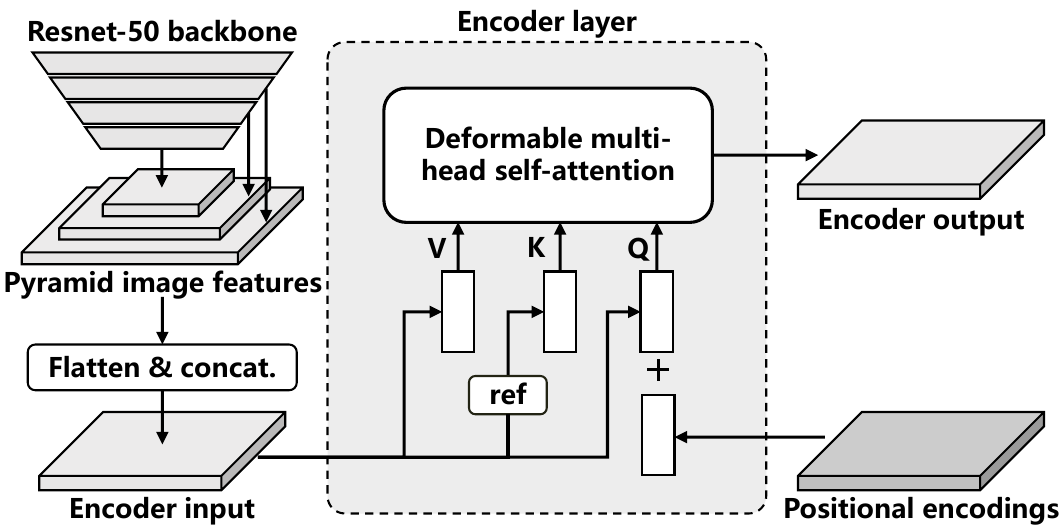}
\caption{
Architecture of the image encoder.
Extracted pyramid image features are flattened, concatenated, and enhanced using deformable self-attention.}
\label{fig:encoder}
\end{figure}

\subsection{Image-Conditioned Anchor Prediction}
\label{subsec:network}

\begin{figure}[t]
\centering
\includegraphics[width=0.9\linewidth]{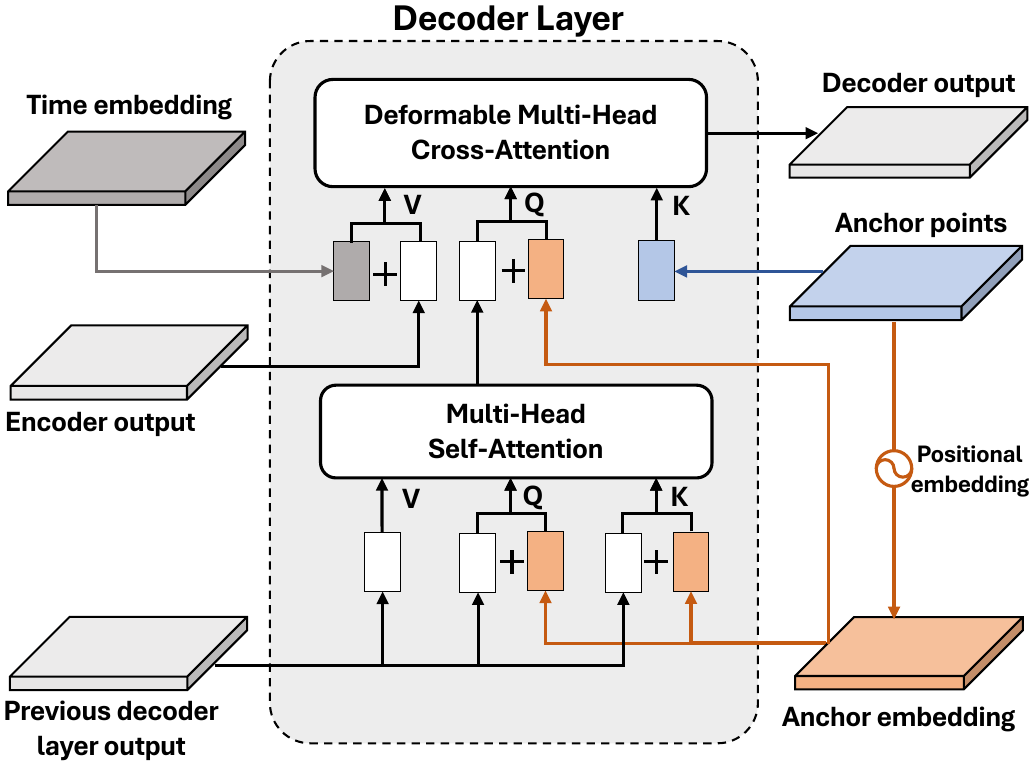}
\caption{
Architecture of the diffusion decoder.
Anchor embeddings serve as spatial queries, while image features and diffusion-timestep embeddings condition anchor interaction and image-to-anchor feature aggregation.
}
\label{fig:decoder}
\end{figure}

The prediction network consists of an image encoder, a diffusion decoder, and two prediction heads.

\paragraph{Image encoder.}
As shown in \Cref{fig:encoder}, a ResNet-50 backbone~\cite{he2016deep} first extracts pyramid image features.
Following A2J-Transformer~\cite{jiang2023a2j}, the multi-scale features are flattened and enhanced using deformable self-attention~\cite{zhu2020deformable}.
After six encoder layers, the image encoder produces the visual feature
\begin{equation}
    \mathbf{F}_{\mathrm{enc}}
    =
    E_{\phi}(\mathbf{I}).
\end{equation}

\paragraph{Diffusion decoder.}
As illustrated in \Cref{fig:decoder}, the coordinates of the input anchors are mapped to anchor embeddings
$\mathbf{E}_{A,t}$,
which serve as spatial queries in the diffusion decoder.
A timestep embedding $\mathbf{E}_t$ is incorporated to indicate the current noise level.

Each decoder layer contains self-attention among anchor queries and deformable cross-attention between anchor queries and multi-scale image features.
The self-attention operation models non-local dependencies among anchors, whereas deformable cross-attention extracts image evidence around the current anchor reference locations.
The decoder output is denoted as
\begin{equation}
    \mathbf{F}_{\mathrm{dec},t}
    =
    D_{\psi}
    \left(
        \mathbf{F}_{\mathrm{enc}},
        \mathbf{E}_{A,t},
        \mathbf{E}_t
    \right)
    \in
    \mathbb{R}^{N_A\times D},
\end{equation}
where $D=256$ in our implementation.

\paragraph{Offset and importance prediction.}
Two multilayer perceptron heads are applied to
$\mathbf{F}_{\mathrm{dec},t}$:
\begin{equation}
    \mathbf{O}_t
    =
    h_{\mathrm{off}}
    \left(
        \mathbf{F}_{\mathrm{dec},t}
    \right),
    \qquad
    \mathbf{W}_t
    =
    h_{\mathrm{imp}}
    \left(
        \mathbf{F}_{\mathrm{dec},t}
    \right).
\end{equation}
Here,
$\mathbf{o}_{t,i,j}$
specifies the 3D offset from anchor $i$ to joint $j$, and
$w_{t,i,j}$
measures the importance of anchor $i$ for predicting joint $j$.

\subsection{Joint-Wise Anchor-to-Joint Ensemble}
\label{subsec:joint_prediction}

Not every dynamic anchor provides reliable evidence for every joint.
Directly aggregating all anchors may therefore introduce proposals with long regression ranges or inaccurate offsets.
We address this issue through joint-wise informative-anchor selection.

For joint $j$, we select the $k$ dynamic anchors with the highest importance logits:
\begin{equation}
    \mathcal{A}_{e,t}(j)
    =
    \operatorname{TopK}_{\mathbf{a}_{t,i}
    \in
    \mathcal{A}_t^{\mathrm{dyn}}}
    \left(
        w_{t,i,j},
        k
    \right).
    \label{eq:ensemble_selection}
\end{equation}
Because the selection is performed independently for each joint,
$\mathcal{A}_{e,t}(j)$
can differ across joints.

The selected dynamic anchors are combined with the fixed context anchors:
\begin{equation}
    \mathcal{S}_t(j)
    =
    \mathcal{A}_{e,t}(j)
    \cup
    \mathcal{A}^{\mathrm{fix}}.
\end{equation}
The importance logits are normalized within
$\mathcal{S}_t(j)$:
\begin{equation}
    \widetilde{w}_{t,i,j}
    =
    \frac{
        \exp(w_{t,i,j})
    }{
        \sum_{\mathbf{a}_{t,r}\in\mathcal{S}_t(j)}
        \exp(w_{t,r,j})
    },
    \qquad
    \mathbf{a}_{t,i}\in\mathcal{S}_t(j).
    \label{eq:normalized_weight}
\end{equation}

The coordinate of joint $j$ at timestep $t$ is then estimated as
\begin{equation}
    \widehat{\mathbf{y}}_{t,j}
    =
    \sum_{\mathbf{a}_{t,i}\in\mathcal{S}_t(j)}
    \widetilde{w}_{t,i,j}
    \left(
        \mathbf{a}_{t,i}
        +
        \mathbf{o}_{t,i,j}
    \right).
    \label{eq:joint_aggregation}
\end{equation}

This ensemble separates anchor exploration from precise localization.
The dynamic-anchor distribution only needs to cover informative neighborhoods, while the residual anchor-to-joint offsets recover the precise joint coordinates.
Moreover, weighted aggregation reduces the influence of perturbations in any individual dynamic anchor.

\begin{figure}[t]
\centering
\includegraphics[width=0.98\linewidth]{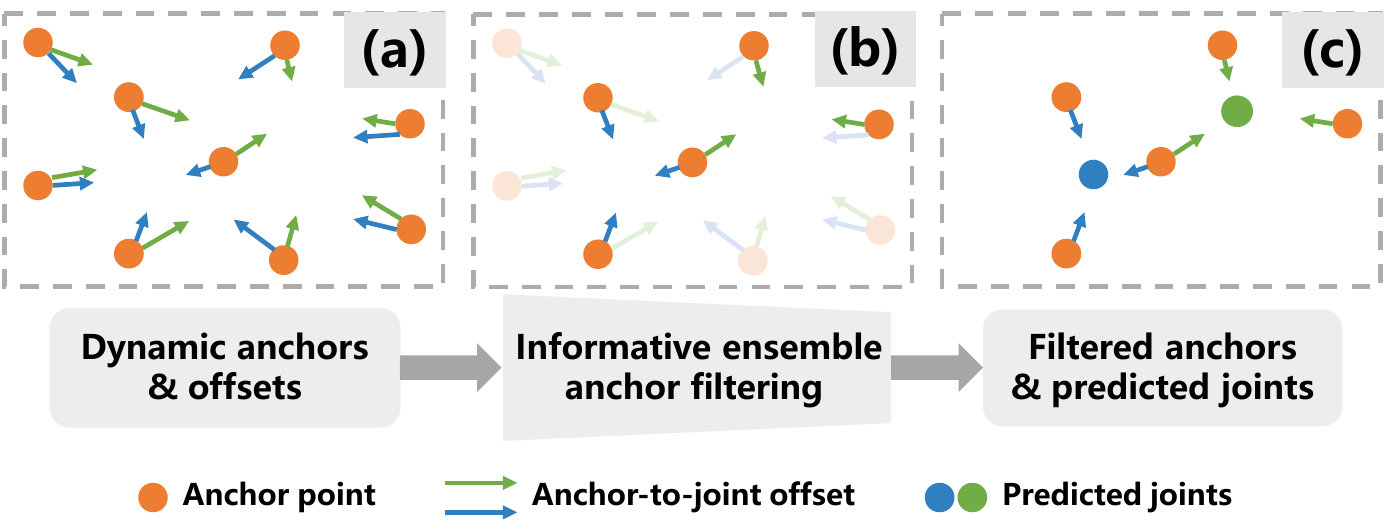}
\caption{
Joint-wise informative-anchor selection and anchor-to-joint aggregation.
For each joint, dynamic anchors with the highest importance scores are selected and combined with the fixed context anchors.
Their joint proposals are then aggregated using normalized importance weights.
}
\label{fig:anchorensemble}
\end{figure}

\subsection{Importance-Guided Anchor Selection and Replenishment}
\label{subsec:anchor_update}

The informative-anchor set in
\Cref{eq:ensemble_selection}
is selected separately for each joint and is used for joint prediction.
For iterative anchor refinement, however, each dynamic anchor must determine the joint region toward which it should move.

Specifically, for the dynamic anchor point $i$, we define its reference parent joint as
\begin{equation}
    p_t(i)
    =
    \arg\max_{j\in\{1,\ldots,N_J\}}
    w_{t,i,j},
    \label{eq:parent_assignment}
\end{equation}
and define its anchor confidence as
\begin{equation}
    c_{t,i}
    =
    \max_{j\in\{1,\ldots,N_J\}}
    w_{t,i,j}.
    \label{eq:anchor_confidence}
\end{equation}

The corresponding denoised-anchor estimate is obtained using the offset toward its parent joint:
\begin{equation}
    \widehat{\mathbf{a}}_{0,i}
    =
    \mathbf{a}_{t,i}^{\mathrm{dyn}}
    +
    \mathbf{o}_{t,i,p_t(i)}.
    \label{eq:denoised_anchor}
\end{equation}
The subscript $0$ indicates that the offset-derived coordinate is treated as an estimate of an informative clean anchor.
Importantly, DiffAnchor does not require
$\widehat{\mathbf{a}}_{0,i}$
to coincide exactly with a joint.
It only needs to lie in a sufficiently informative neighborhood from which a short-range offset can accurately recover the target coordinate.

Using
$\widehat{\mathbf{A}}_0^{\mathrm{dyn}}$
as the clean-anchor estimate, DDIM~\cite{song2020denoising} produces the candidate dynamic anchors at the next selected timestep.
For deterministic DDIM sampling, the implied noise estimate is
\begin{equation}
    \widehat{\boldsymbol{\epsilon}}_t
    =
    \frac{
        \mathbf{A}_t^{\mathrm{dyn}}
        -
        \sqrt{\bar{\alpha}_t}
        \widehat{\mathbf{A}}_0^{\mathrm{dyn}}
    }{
        \sqrt{1-\bar{\alpha}_t}
    },
    \label{eq:estimated_noise}
\end{equation}
and the candidate anchors at timestep $t'$ are obtained by
\begin{equation}
    \widetilde{\mathbf{A}}_{t'}^{\mathrm{dyn}}
    =
    \sqrt{\bar{\alpha}_{t'}}
    \widehat{\mathbf{A}}_0^{\mathrm{dyn}}
    +
    \sqrt{1-\bar{\alpha}_{t'}}
    \widehat{\boldsymbol{\epsilon}}_t,
    \qquad
    t'<t.
    \label{eq:ddim_update}
\end{equation}

\begin{figure}[t]
\centering
\includegraphics[width=0.98\linewidth]{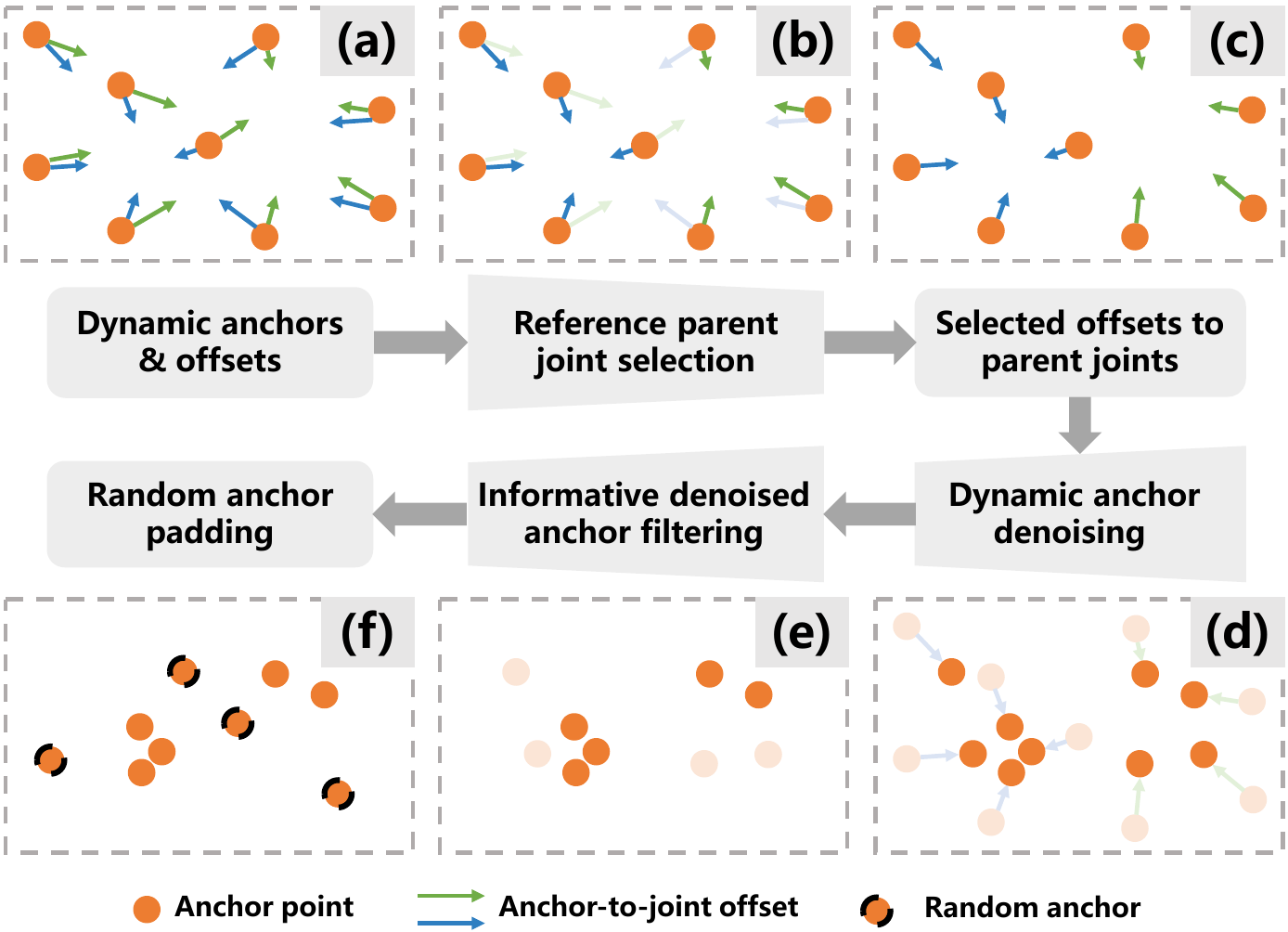}
\caption{
Importance-guided dynamic-anchor denoising and updating.
Each dynamic anchor is assigned to its most confident parent joint and moved according to the corresponding anchor-to-joint offset.
High-confidence denoised anchors are retained, whereas the remaining anchors are replaced by newly sampled candidates.
}
\label{fig:anchordiffusion}
\end{figure}

Some denoised anchors rapidly approach useful joint regions, whereas others remain unreliable.
To retain useful discoveries while preserving exploration, we select the $m$ dynamic anchors with the highest confidence:
\begin{equation}
    \mathcal{A}_{d,t'}
    =
    \operatorname{TopK}_{i}
    \left(
        c_{t,i},
        m
    \right)
    \subset
    \widetilde{\mathcal{A}}_{t'}^{\mathrm{dyn}}.
    \label{eq:retained_anchors}
\end{equation}

The remaining
$N_{\mathrm{dyn}}-m$
positions are replenished with newly sampled Gaussian anchors:
\begin{equation}
    \mathcal{A}_{t'}^{\mathrm{dyn}}
    =
    \mathcal{A}_{d,t'}
    \cup
    \mathcal{A}_{t'}^{\mathrm{new}},
    \qquad
    \left|
        \mathcal{A}_{t'}^{\mathrm{new}}
    \right|
    =
    N_{\mathrm{dyn}}-m.
    \label{eq:anchor_replenishment}
\end{equation}
The newly sampled anchors follow the noise level associated with timestep $t'$.

The retained anchors exploit joint regions that have already been identified, while the replenished anchors continue to explore insufficiently covered regions.
The updated dynamic anchors are concatenated with the unchanged fixed anchors and passed to the next denoising step.
After $S$ sampling steps, joint prediction of the last step is used as the final output:
\begin{equation}
    \widehat{\mathbf{Y}}
    =
    \widehat{\mathbf{Y}}_{0}.
\end{equation}

\subsection{Training Objectives}
\label{subsec:training_objectives}

DiffAnchor is trained using a joint estimation loss, a denoised-anchor loss, and an anchor surrounding loss:
\begin{equation}
    \mathcal{L}
    =
    \lambda_J\mathcal{L}_J
    +
    \lambda_D\mathcal{L}_D
    +
    \lambda_A\mathcal{L}_A.
    \label{eq:total_loss}
\end{equation}

We first define the robust regression function
\begin{equation}
    \rho_{\tau}(x)
    =
    \begin{cases}
        \dfrac{x^2}{2\tau},
        &
        |x|<\tau,
        \\[5pt]
        |x|-\dfrac{\tau}{2},
        &
        \text{otherwise}.
    \end{cases}
    \label{eq:robust_loss}
\end{equation}

\paragraph{Joint estimation loss.}
The joint estimation loss directly supervises the aggregated pose prediction.
Following anchor-based regression methods~\cite{xiong2019a2j,jiang2023a2j}, we separately handle the in-plane and depth coordinates:
\begin{equation}
\begin{aligned}
    \mathcal{L}_J
    =
    \frac{1}{N_J}
    \sum_{j=1}^{N_J}
    \bigg[
    &
    \alpha
    \sum_{c\in\{x,y\}}
    \rho_{\tau_{xy}}
    \left(
        \widehat{y}_{t,j}^{c}
        -
        y_j^{c}
    \right)
    +
    \rho_{\tau_z}
    \left(
        \widehat{y}_{t,j}^{z}
        -
        y_j^{z}
    \right)
    \bigg].
    \label{eq:joint_loss}
\end{aligned}
\end{equation}
We set
$\alpha=0.5$,
$\tau_{xy}=1$,
and
$\tau_z=3$.

\paragraph{Denoised-anchor loss.}
The denoised-anchor loss encourages each dynamic anchor to move toward its assigned parent-joint region:
\begin{equation}
    \mathcal{L}_D
    =
    \frac{1}{N_{\mathrm{dyn}}}
    \sum_{i=1}^{N_{\mathrm{dyn}}}
    \rho_{\tau_D}
    \left(
        \widehat{\mathbf{a}}_{0,i}
        -
        \mathbf{y}_{p_t(i)}
    \right).
    \label{eq:denoised_anchor_loss}
\end{equation}
We use $\tau_D=100$.
The relatively large transition threshold avoids forcing every anchor to coincide exactly with a joint and instead encourages it to enter a sufficiently informative local neighborhood.

\paragraph{Anchor surrounding loss.}
In addition to supervising individual denoised anchors, we encourage the weighted centers of the dynamic-anchor distribution to surround each target joint.
We first normalize the importance logits over all dynamic anchors:
\begin{equation}
    \overline{w}_{t,i,j}
    =
    \frac{
        \exp(w_{t,i,j})
    }{
        \sum_{r=1}^{N_{\mathrm{dyn}}}
        \exp(w_{t,r,j})
    }.
\end{equation}
The corresponding global dynamic-anchor center for joint $j$ is
\begin{equation}
    \overline{\mathbf{a}}_{t,j}^{\mathrm{dyn}}
    =
    \sum_{i=1}^{N_{\mathrm{dyn}}}
    \overline{w}_{t,i,j}
    \mathbf{a}_{t,i}^{\mathrm{dyn}}.
\end{equation}

We additionally compute the weighted center of the selected informative anchors:
\begin{equation}
    \overline{\mathbf{a}}_{t,j}^{e}
    =
    \sum_{\mathbf{a}_{t,i}
    \in
    \mathcal{A}_{e,t}(j)}
    \widetilde{w}_{t,i,j}^{e}
    \mathbf{a}_{t,i},
\end{equation}
where
$\widetilde{w}_{t,i,j}^{e}$
is obtained by applying softmax within
$\mathcal{A}_{e,t}(j)$.

The anchor surrounding loss is defined as
\begin{equation}
\begin{aligned}
    \mathcal{L}_A
    =
    \frac{1}{N_J}
    \sum_{j=1}^{N_J}
    \bigg[
    &
    \rho_{\tau_A}
    \left(
        \overline{\mathbf{a}}_{t,j}^{\mathrm{dyn}}
        -
        \mathbf{y}_j
    \right)
    +
    \beta
    \rho_{\tau_A}
    \left(
        \overline{\mathbf{a}}_{t,j}^{e}
        -
        \mathbf{y}_j
    \right)
    \bigg].
    \label{eq:anchor_surrounding_loss}
\end{aligned}
\end{equation}
We set $\tau_A=1$ and $\beta=1.5$, assigning larger importance to the selected anchors that directly participate in ensemble.

The final loss weights are set to
\begin{equation}
    \lambda_J=3,
    \qquad
    \lambda_D=1,
    \qquad
    \lambda_A=1.
\end{equation}

\begin{table}[t]
\centering
\small
\setlength{\tabcolsep}{5pt}
\caption{
Comparison with state-of-the-art methods on HO-3D V2.
MPJPE is calculated after scale and translation alignment.
}
\begin{tabular}{lcc}
\toprule
Methods
& MPJPE (mm) $\downarrow$
& Mean joint AUC $\uparrow$ \\
\midrule
\multicolumn{3}{c}{Model-based} \\
\midrule
I2L-MeshNet~\cite{moon2020i2l}
& 26.0 & 52.9 \\
ArtiBoost~\cite{yang2022artiboost}
& 25.3 & 53.2 \\
HandOccNet~\cite{park2022handoccnet}
& 24.0 & 55.7 \\
Wang et al.~\cite{wang2023interacting}
& 23.8 & 56.0 \\
HOISDF~\cite{qi2024hoisdf}
& 22.8 & -- \\
NVF~\cite{huang2023nvf}
& 21.8 & 61.0 \\
\midrule
\multicolumn{3}{c}{Model-free} \\
\midrule
KPT~\cite{hampali2022keypoint}
& 25.7 & 55.3 \\
A2J-Transformer~\cite{jiang2023a2j}
& 24.0 & 58.9 \\
A2J-Transformer+~\cite{jiang20253d}
& 20.0 & 62.7 \\
\rowcolor{Gray}
\textbf{DiffAnchor (Ours)}
& \textbf{18.5} & \textbf{64.8} \\
\bottomrule
\end{tabular}
\label{tab:HO_3D}
\end{table}

\begin{table}[t]
\small
\centering
\setlength{\tabcolsep}{8pt}
\caption{
Comparison with state-of-the-art methods on NYU~\cite{tompson2014real} and ICVL~\cite{tang2014latent}.
``P'', ``V'', and ``D'' denote point-cloud, voxel, and depth-image inputs, respectively.
}
\begin{tabular}{lccc}
\toprule
\multirow{2}{*}{Methods}
& \multicolumn{2}{c}{Mean joint error (mm) $\downarrow$}
& \multirow{2}{*}{Input} \\
\cline{2-3}
\addlinespace[1.2pt]
& NYU & ICVL & \\
\midrule
HandPointNet~\cite{ge2018hand}
& 10.54 & 6.94 & P \\
P2P~\cite{ge2018point}
& 9.10 & 6.30 & P \\
V2V-PoseNet~\cite{moon2018v2v}
& 8.42 & 6.28 & V \\
HandFoldingNet~\cite{cheng2021handfoldingnet}
& 8.58 & 5.95 & P \\
CrossInfoNet~\cite{du2019crossinfonet}
& 10.08 & 6.73 & D \\
A2J~\cite{xiong2019a2j}
& 8.61 & 6.46 & D \\
JGR-P2O~\cite{fang2020jgr}
& 8.29 & 6.02 & D \\
VVS~\cite{cheng2022efficient}
& 7.79 & 6.22 & D \\
AWR~\cite{huang2020awr}
& 7.48 & 5.98 & D \\
A2J-Transformer~\cite{jiang2023a2j}
& 7.88 & 6.06 & D \\
DenseRecurrent~\cite{deng2022recurrent}
& 6.85 & 6.05 & D \\
DF-Mamba~\cite{zhou2026df}
& 7.96 & -- & D \\
HandDiff~\cite{cheng2024handdiff}
& 7.38 & 5.72 & D+P \\
A2J-Transformer+~\cite{jiang20253d}
& 7.37 & 5.81 & D \\
\rowcolor{Gray}
\textbf{DiffAnchor (Ours)}
& \textbf{6.81} & \textbf{5.59} & D \\
\bottomrule
\end{tabular}
\label{tab:nyu_icvl}
\end{table}

\section{Experiments}
\label{sec:experiment}

\subsection{Datasets and Evaluation Metrics}
\label{subsec:datasets}

We evaluate DiffAnchor on four widely used 3D hand pose estimation benchmarks. HO-3D V2~\cite{hampali2020honnotate} contains about 66K/11K RGB training/test images with hand--object interaction; following~\cite{hampali2022keypoint}, we report scale- and translation-aligned MPJPE and mean joint AUC. NYU~\cite{tompson2014real} contains about 72K/8.2K depth images and is evaluated on the standard 14-joint subset using mean 3D joint error. ICVL~\cite{tang2014latent} contains about 22K/1.5K depth images and uses the same mean-error metric. InterHand2.6M~\cite{moon2020interhand2} contains single and interacting hand RGB images. We follow the baseline~\cite{moon2020interhand2} protocol and report root-aligned MPJPE for single, two, and all hand subsets.

\begin{table}[t]
\centering
\small
\setlength{\tabcolsep}{5pt}
\caption{
Comparison with state-of-the-art methods on InterHand2.6M.
Results marked with $\dagger$ use model-based evaluation settings, while the remaining methods follow the model-free protocol of InterNet~\cite{moon2020interhand2}.
}
\begin{tabular}{lccc}
\toprule
\multirow{2}{*}{Methods}
& \multicolumn{3}{c}{MPJPE (mm) $\downarrow$} \\
\cline{2-4}
\addlinespace[1.2pt]
& Single hand & Two hands & All hands \\
\midrule
\multicolumn{4}{c}{Model-based} \\
\midrule
$\dagger$ InterShape~\cite{zhang2021interacting}
& -- & 13.07 & -- \\
$\dagger$ IntagHand~\cite{li2022interacting}
& -- & 10.30 & -- \\
$\dagger$ RenderIH~\cite{li2023renderih}
& -- & 10.66 & -- \\
$\dagger$ DIR~\cite{ren2023decoupled}
& -- & 8.70 & -- \\
\midrule
\multicolumn{4}{c}{Model-free} \\
\midrule
InterNet~\cite{moon2020interhand2}
& 12.16 & 16.02 & 14.22 \\
DIGIT~\cite{fan2021learning}
& 11.32 & 15.57 & -- \\
KPT~\cite{hampali2022keypoint}
& 10.99 & 14.34 & 12.78 \\
A2J-Transformer~\cite{jiang2023a2j}
& 8.10 & 10.96 & 9.63 \\
DF-Mamba~\cite{zhou2026df}
& 7.94 & 10.53 & 9.32 \\
A2J-Transformer+~\cite{jiang20253d}
& \textbf{7.58} & 10.45 & \textbf{9.11} \\
\rowcolor{Gray}
\textbf{DiffAnchor (Ours)}
& 7.85 & \textbf{10.36} & 9.18 \\
\bottomrule
\end{tabular}
\label{tab:interhand}
\end{table}

\subsection{Implementation Details}
\label{subsec:implementation}

DiffAnchor is implemented in PyTorch and trained on one RTX 3090. RGB and depth crops are resized to $256\times256$ and $176\times176$, respectively, following the augmentation protocols of the corresponding dataset baselines. We train for 62 epochs with AdamW, learning rate and weight decay of $10^{-4}$, and use $N_A=800$ ($N_{\mathrm{dyn}}=544$, $N_{\mathrm{fix}}=256$) anchors. The forward process uses 1000 timesteps, and inference uses 3 deterministic DDIM steps and retains $m=300$ dynamic anchors.



\begin{table}[t]
\centering
\small
\setlength{\tabcolsep}{12pt}
\caption{
Ablation of the main components of DiffAnchor on InterHand2.6M.
}
\begin{tabular}{lc}
\toprule
Component
& MPJPE (mm) $\downarrow$ \\
\midrule
w/o dynamic anchors $\mathcal{A}^{\mathrm{dyn}}$
& 9.42 \\
w/o fixed context anchors $\mathcal{A}^{\mathrm{fix}}$
& 9.34 \\
w/o informative ensemble anchors $\mathcal{A}_{e}$
& 9.39 \\
w/o retained denoised anchors $\mathcal{A}_{d}$
& 9.35 \\
w/o denoised-anchor loss $\mathcal{L}_{D}$
& 9.40 \\
\midrule
Full model
& \textbf{9.18} \\
\bottomrule
\end{tabular}
\label{tab:ablation_component}
\end{table}

\begin{table}[t]
\small
\centering
\setlength{\tabcolsep}{6pt}
\caption{
Effects of the number of DDIM sampling steps and retained denoised anchors.
}
\begin{tabular}{ccccc}
\toprule
\multirow{2}{*}{Retained anchors}
& \multicolumn{4}{c}{MPJPE (mm) $\downarrow$} \\
\cline{2-5}
\addlinespace[1pt]
& Step 1 & Step 2 & Step 3 & Step 5 \\
\midrule
0
& 11.80 & 9.42 & 9.44 & 9.42 \\
150
& 11.79 & 9.24 & 9.20 & 9.19 \\
300
& \textbf{11.78}
& \textbf{9.21}
& \textbf{9.18}
& \textbf{9.18} \\
500
& 11.79 & 9.24 & 9.21 & 9.21 \\
\bottomrule
\end{tabular}
\label{tab:ablation_inference}
\end{table}

\begin{table}[t]
\small
\centering
\setlength{\tabcolsep}{10pt}
\caption{
Comparison of dynamic-anchor construction strategies during training.
}
\begin{tabular}{lc}
\toprule
Dynamic-anchor construction
& MPJPE (mm) $\downarrow$ \\
\midrule
Repeated GT anchors
& 9.23 \\
Noisy repeated GT anchors
& 9.20 \\
GT joints + Gaussian anchors
& \textbf{9.18} \\
\bottomrule
\end{tabular}
\label{tab:anchor_generation}
\end{table}

\subsection{Comparison with State-of-the-Art Methods}
\label{subsec:sota}

\textbf{HO-3D V2.}
As shown in \Cref{tab:HO_3D}, DiffAnchor achieves the best performance under both evaluation metrics.
Compared with the strongest previous model-free method, A2J-Transformer+, DiffAnchor reduces MPJPE from 20.0 mm to 18.5 mm, corresponding to an improvement of 1.5 mm or 7.5\%.
It also increases the mean joint AUC from 62.7 to 64.8, yielding a gain of 2.1 percentage points.
These results are particularly relevant to embodied interaction because HO-3D V2 contains severe hand--object occlusions, for which informative local regressors cannot be reliably covered by a fixed anchor distribution.
The clear improvement supports the effectiveness of dynamically exploring and retaining image-dependent anchor regions.

\textbf{NYU and ICVL.}
The results in \Cref{tab:nyu_icvl} demonstrate that DiffAnchor generalizes effectively from RGB images to depth inputs without modifying its main architecture.
On NYU, DiffAnchor obtains a mean joint error of 6.81 mm, outperforming the previous best result of 6.85 mm.
On ICVL, it achieves the best listed result of 5.59~mm, outperforming HandDiff and A2J-Transformer+ by 0.13 and 0.22~mm, respectively. 
The improvements on both depth benchmarks indicate that anchor-space diffusion is not tied to a particular image modality.

\textbf{InterHand2.6M.}
As reported in \Cref{tab:interhand}, DiffAnchor achieves the best model-free performance for interacting hands, with an MPJPE of 10.36 mm.
It improves upon A2J-Transformer+ by 0.09 mm and A2J-Transformer by 0.60 mm.
For the single-hand and all-hand settings, DiffAnchor obtains 7.85 mm and 9.18 mm, respectively, which are competitive with the strongest deterministic baseline.
The larger improvement in the two-hand setting suggests that dynamically exploring local anchor regions is particularly helpful when severe hand--hand occlusions make fixed anchors less informative.

\subsection{Ablation Studies}
\label{subsec:ablation}

Unless otherwise specified, all ablation experiments are performed using the base DiffAnchor model on InterHand2.6M.

\subsubsection{Component Analysis}

The results are summarized in \Cref{tab:ablation_component}. The full model obtains 9.18~mm. Removing dynamic anchors causes the largest degradation (+0.24~mm), confirming the value of image-adaptive local regressors. Removing joint-wise informative selection or the denoised-anchor loss increases error by 0.21 and 0.22~mm, respectively, while removing fixed anchors or retained denoised anchors costs 0.16 and 0.17~mm. Thus, adaptive exploration, stable global references, selective aggregation, and retention are complementary.






\subsubsection{Inference Strategy}

\Cref{tab:ablation_inference} shows that most improvement occurs by the second step and performance saturates at three steps. Retaining anchors improves over full resampling, with $m=300$ giving the best trade-off; retaining 500 anchors slightly degrades accuracy because more unreliable anchors survive. We therefore use three steps and $m=300$.



\subsubsection{Training Anchor Construction}

During training, the ground-truth joints occupy part of the clean dynamic-anchor set, and additional anchors are required to reach the predefined number $N_{\mathrm{dyn}}$.
We compare three padding strategies:
repeating the ground-truth joints,
adding small perturbations to the repeated joints,
and sampling the remaining anchors from a Gaussian distribution.

As shown in \Cref{tab:anchor_generation}, the three strategies yield relatively similar performance, indicating that DiffAnchor is not highly sensitive to the exact padding mechanism.
Nevertheless, combining the ground-truth joints with Gaussian anchors achieves the best result.
The Gaussian anchors expose the model to a broader initial spatial distribution and are therefore adopted as the default training strategy.

\begin{table}[t]
\centering
\small
\setlength{\tabcolsep}{4pt}
\caption{
Computational cost and performance comparison on InterHand2.6M.
The base and small variants use six and three diffusion-decoder layers, respectively.
GPU memory is reported in MB.
}
\begin{tabular}{lcccc}
\toprule
Methods
& FPS $\uparrow$
& Params. $\downarrow$
& Memory $\downarrow$
& MPJPE $\downarrow$ \\
\midrule
A2J-Transformer
& 25.65
& 41.89 M
& 698.05
& 9.63 \\
\rowcolor{Gray}
DiffAnchor (Base)
& 15.13
& 43.53 M
& 725.63
& \textbf{9.18} \\
\rowcolor{Gray}
DiffAnchor (Small)
& 25.61
& \textbf{40.33 M}
& \textbf{661.94}
& 9.20 \\
\bottomrule
\end{tabular}
\label{tab:computational_cost}
\end{table}

\subsection{Efficiency and Stability Analysis}
\label{subsec:efficiency}

\subsubsection{Computational Cost}

The base and small variants use 6 and 3 decoder layers. As shown in \Cref{tab:computational_cost}, the small model matches A2J-Transformer throughput, uses fewer parameters and memory, and improves MPJPE from 9.63 to 9.20~mm. Relative to the base variant, it loses only 0.02~mm while increasing speed from 15.13 to 25.61 FPS.



\subsubsection{Random-Seed Stability}

\begin{figure}[t]
\centering
\includegraphics[width=0.95\linewidth]{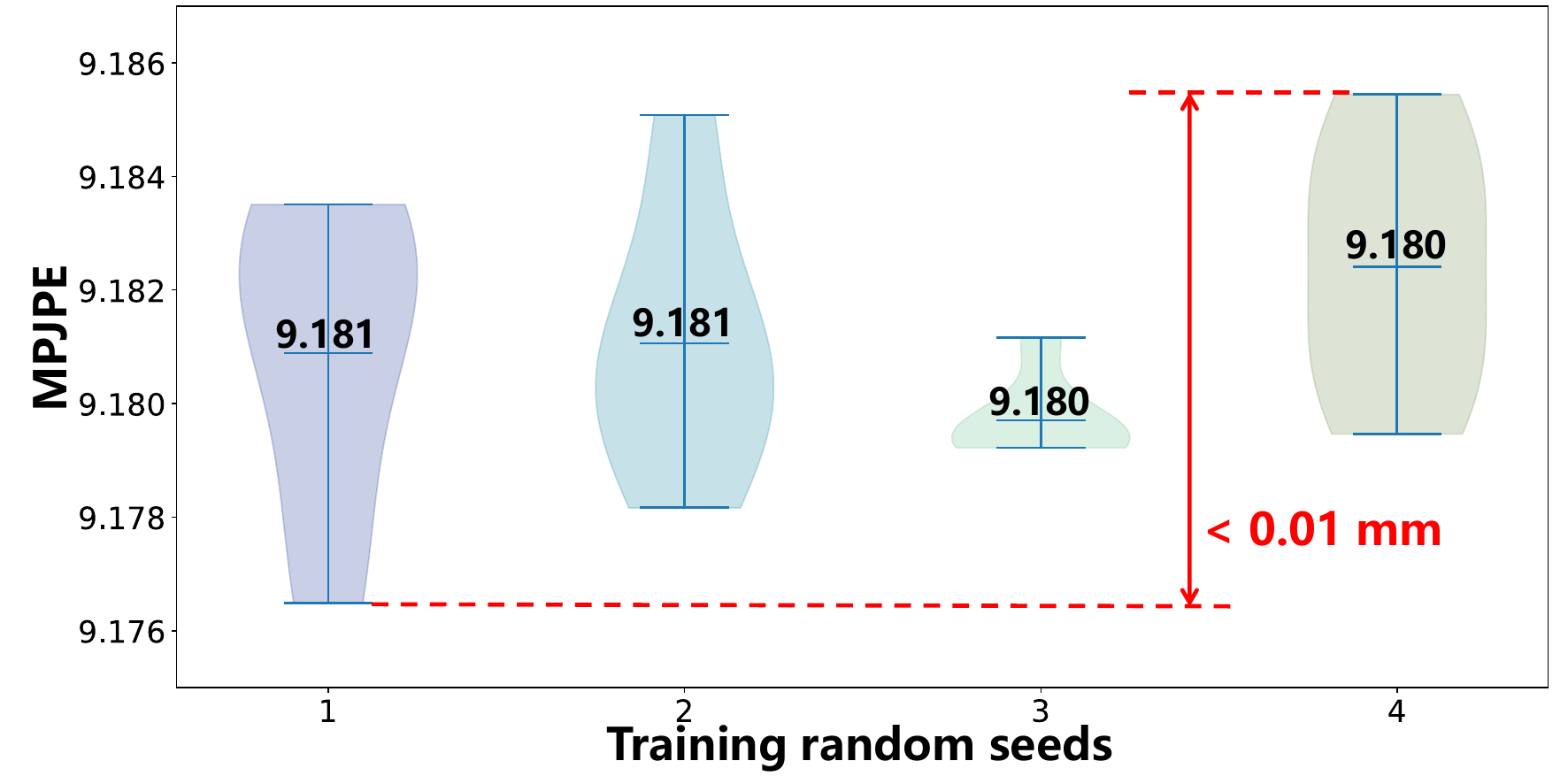}
\vspace{-8pt}
\caption{
Random-seed stability analysis.
Four models are independently trained using different random seeds, and each model is evaluated ten times using different inference seeds.
}
\label{fig:violin}
\end{figure}

\begin{table}[t]
\centering
\small
\setlength{\tabcolsep}{8pt}
\caption{
Joint-wise comparison with A2J-Trans.
The anatomical order from the root toward the fingertip is MCP, PIP, DIP, and TIP.
}
\begin{tabular}{cccc}
\toprule
\multirow{2}{*}{Joint type}
& \multicolumn{3}{c}{MPJPE (mm) $\downarrow$} \\
\cline{2-4}
\addlinespace[1.2pt]
& A2J-Trans.
& DiffAnchor
& Improvement \\
\midrule
TIP
& 13.12 & 12.46
& \textbf{0.66} / \textbf{5.04\%} \\
DIP
& 10.70 & 10.19
& 0.51 / 4.77\% \\
PIP
& 9.17 & 8.74
& 0.43 / 4.69\% \\
MCP
& 7.47 & 7.19
& 0.28 / 3.72\% \\
\bottomrule
\end{tabular}
\label{tab:joint_wise_improvement}
\end{table}

\begin{figure}[t]
\centering
\includegraphics[width=0.96\linewidth]{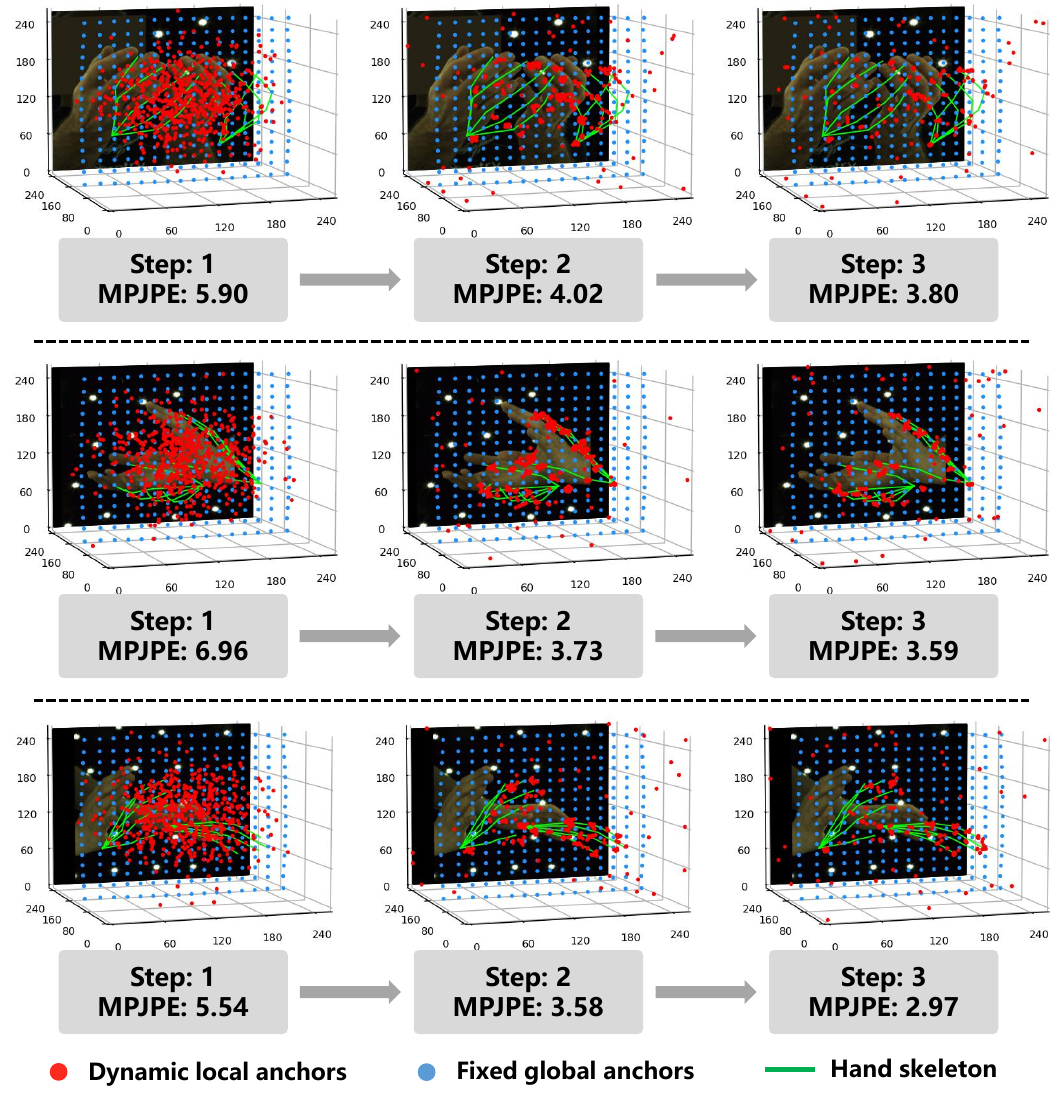}
\caption{
Examples of dynamic-anchor movement and the corresponding pose-estimation errors over three sampling steps on InterHand2.6M.
}
\label{fig:vis_anchor_movement}
\end{figure}

DiffAnchor initializes dynamic anchors stochastically, which may raise concerns about sensitivity to random seeds.
We therefore train four models using different random seeds and evaluate each model ten times with independently sampled inference anchors.
As shown in \Cref{fig:violin}, the mean MPJPE of all four models remains close to 9.18 mm.
The corresponding standard deviations are 0.0014, 0.00012, 0.0005, and 0.0011 mm, respectively.
The extremely small variations indicate that the weighted anchor-to-joint ensemble and fixed context anchors effectively suppress the influence of individual random initializations.

\subsection{Further Analysis and Qualitative Results}
\label{subsec:further_analysis}

\subsubsection{Joint-Wise Improvement}

To examine where dynamic anchors provide the largest benefit, we compare the errors of different joint types with A2J-Transformer.
As shown in \Cref{tab:joint_wise_improvement}, DiffAnchor improves all four joint groups.
Moreover, the gain generally increases with the distance from the hand root.
The largest improvement occurs for fingertip joints, whose MPJPE is reduced by 0.66 mm or 5.04\%.

Distal joints are typically more difficult to localize because they exhibit larger motion ranges and are more frequently affected by self-occlusion and interaction occlusion.
The stronger improvement on these joints supports the motivation of DiffAnchor: dynamically generated anchors can move toward difficult local regions and reduce the regression range required for precise fingertip localization.

\subsubsection{Anchor Evolution}


\Cref{fig:vis_anchor_movement} visualizes the evolution of the dynamic anchors over 3 sampling steps.
Starting from a random distributed initialization, the dynamic anchors progressively approach local hand-joint regions.
The largest spatial adjustment and error reduction occur during the first two steps, while the third step further refines the anchor distribution and produces near-saturated predictions.
These observations are consistent with the quantitative step analysis in \Cref{tab:ablation_inference}.

The visualizations also show that prediction accuracy improves as informative anchors become more densely distributed around the hand structure.
This supports our interpretation that diffusion is not required to recover the exact joint coordinates directly.
Instead, it identifies useful local regression regions, after which anchor-to-joint offsets complete precise localization.

\subsubsection{Qualitative Results and Limitations}

\begin{figure}[t]
\centering
\includegraphics[width=0.9\linewidth]{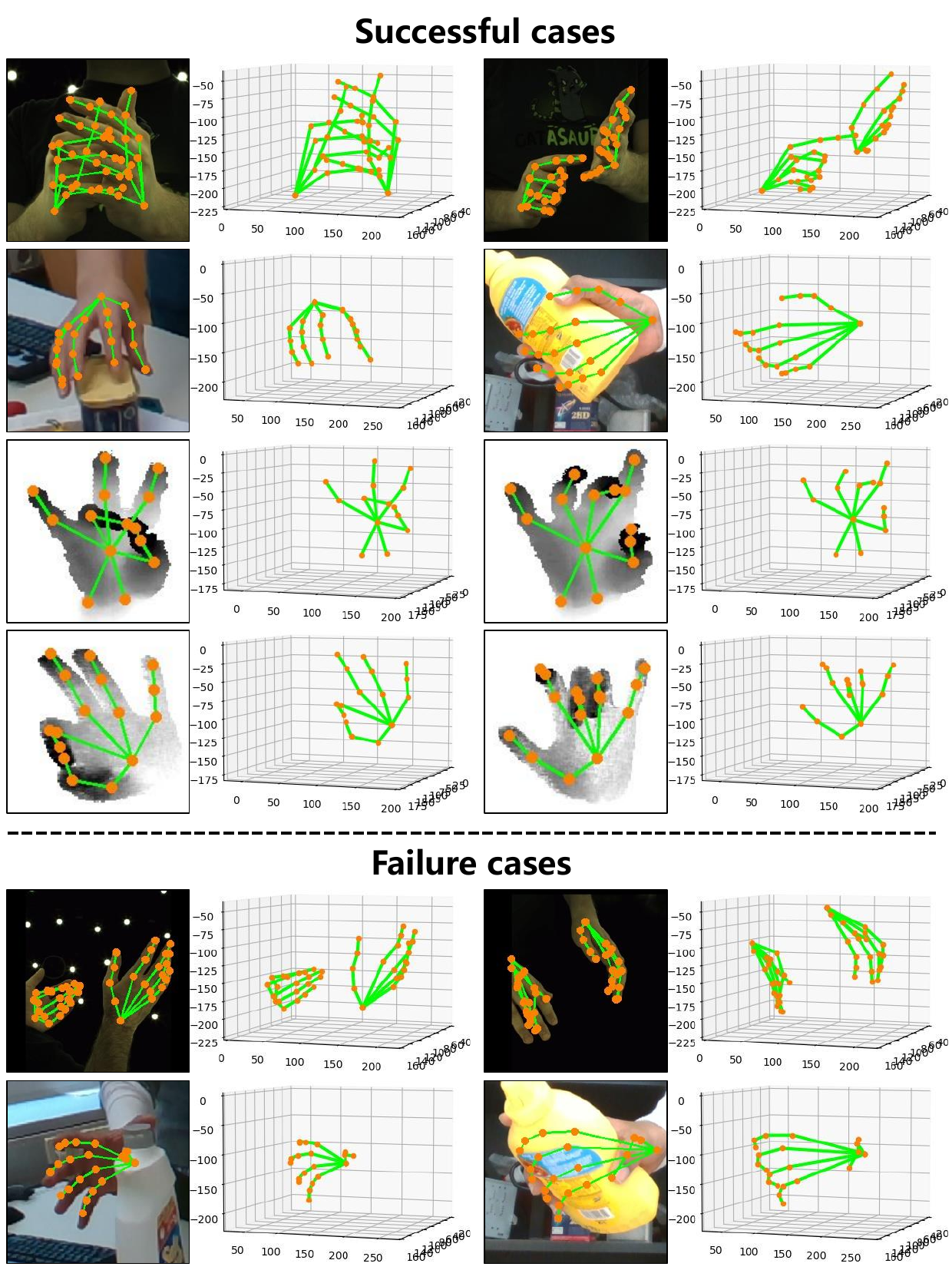}
\caption{
Qualitative results and representative failure cases on HO-3D V2, NYU, ICVL, and InterHand2.6M.
}
\label{fig:qualitative}
\end{figure}

As shown in \Cref{fig:qualitative}, DiffAnchor produces accurate 3D hand poses across hand--object interaction, single-hand, and interacting-hand scenarios, as well as across RGB and depth inputs.
The method remains effective under moderate self-occlusion and interaction occlusion because the dynamic anchors can adapt their spatial distribution to the available image evidence.

The main failure cases occur when a large portion of the hand is severely occluded or absent from the observation.
Under such conditions, the image provides insufficient evidence for assigning reliable importance scores and anchor-to-joint offsets.
The replenished anchors can continue exploring alternative regions, but they cannot fully resolve ambiguity when the relevant visual evidence is entirely missing.

\section{Conclusion}
In this article, we introduce DiffAnchor, an anchor-space diffusion framework for monocular 3D hand pose estimation. 
DiffAnchor demonstrates that diffusion models can be effectively utilized by assigning stochastic exploration to an intermediate anchor space while preserving precise localization through anchor-to-joint regression.
This approach allows anchors to dynamically adapt based on the input hand pattern, thereby simplifying the prediction of accurate anchor-to-joint offsets.
Furthermore, we propose a novel method for anchor selection and updating, which helps the model efficiently acquire informative anchors.
Extensive experiments validate the superior performance and generalizability of DiffAnchor across diverse application scenarios and input modalities, demonstrating significant improvements over previous methods while maintaining a comparable computational cost.

\bibliographystyle{IEEEtran}
\bibliography{main}

@String(CVPR  = {IEEE Conf. Comput. Vis. Pattern Recog.})

@String(AAAI  = {AAAI})

@String(TOG   = {ACM Trans. Graph.})

@String(CVPR  = {CVPR})

@String(TOG   = {ACM TOG})

@inproceedings{moon2020interhand2,
  title={Interhand2. 6m: A dataset and baseline for 3d interacting hand pose estimation from a single rgb image},
  author={Moon, Gyeongsik and Yu, Shoou-I and Wen, He and Shiratori, Takaaki and Lee, Kyoung Mu},
  booktitle={Proceedings of the IEEE/CVF European Conference on Computer Vision},
  pages={548--564},
  year={2020},
  organization={Springer}
}

@inproceedings{jiang2023a2j,
  title={A2J-Transformer: Anchor-to-Joint Transformer Network for 3D Interacting Hand Pose Estimation from a Single RGB Image},
  author={Jiang, Changlong and Xiao, Yang and Wu, Cunlin and Zhang, Mingyang and Zheng, Jinghong and Cao, Zhiguo and Zhou, Joey Tianyi},
  booktitle={Proceedings of the IEEE/CVF Conference on Computer Vision and Pattern Recognition},
  pages={8846--8855},
  year={2023}
}

@inproceedings{xiong2019a2j,
  title={A2j: Anchor-to-joint regression network for 3d articulated pose estimation from a single depth image},
  author={Xiong, Fu and Zhang, Boshen and Xiao, Yang and Cao, Zhiguo and Yu, Taidong and Zhou, Joey Tianyi and Yuan, Junsong},
  booktitle={Proceedings of the IEEE/CVF International Conference on Computer Vision},
  pages={793--802},
  year={2019}
}

@inproceedings{hampali2022keypoint,
  title={Keypoint transformer: Solving joint identification in challenging hands and object interactions for accurate 3d pose estimation},
  author={Hampali, Shreyas and Sarkar, Sayan Deb and Rad, Mahdi and Lepetit, Vincent},
  booktitle={Proceedings of the IEEE/CVF Conference on Computer Vision and Pattern Recognition},
  pages={11090--11100},
  year={2022}
}

@inproceedings{fang2020jgr,
  title={Jgr-p2o: Joint graph reasoning based pixel-to-offset prediction network for 3d hand pose estimation from a single depth image},
  author={Fang, Linpu and Liu, Xingyan and Liu, Li and Xu, Hang and Kang, Wenxiong},
  booktitle={Proceedings of the European Conference on Computer Vision},
  pages={120--137},
  year={2020},
  organization={Springer}
}

@inproceedings{fan2021learning,
  title={Learning to disambiguate strongly interacting hands via probabilistic per-pixel part segmentation},
  author={Fan, Zicong and Spurr, Adrian and Kocabas, Muhammed and Tang, Siyu and Black, Michael J and Hilliges, Otmar},
  booktitle={2021 International Conference on 3D Vision (3DV)},
  pages={1--10},
  year={2021},
  organization={IEEE}
}

@inproceedings{zhang2021interacting,
  title={Interacting two-hand 3d pose and shape reconstruction from single color image},
  author={Zhang, Baowen and Wang, Yangang and Deng, Xiaoming and Zhang, Yinda and Tan, Ping and Ma, Cuixia and Wang, Hongan},
  booktitle={Proceedings of the IEEE/CVF International Conference on Computer Vision},
  pages={11354--11363},
  year={2021}
}

@inproceedings{meng20223d,
  title={3d interacting hand pose estimation by hand de-occlusion and removal},
  author={Meng, Hao and Jin, Sheng and Liu, Wentao and Qian, Chen and Lin, Mengxiang and Ouyang, Wanli and Luo, Ping},
  booktitle={Proceedings of the European Conference on Computer Vision},
  pages={380--397},
  year={2022},
  organization={Springer}
}

@inproceedings{li2022interacting,
  title={Interacting attention graph for single image two-hand reconstruction},
  author={Li, Mengcheng and An, Liang and Zhang, Hongwen and Wu, Lianpeng and Chen, Feng and Yu, Tao and Liu, Yebin},
  booktitle={Proceedings of the IEEE/CVF Conference on Computer Vision and Pattern Recognition},
  pages={2761--2770},
  year={2022}
}

@inproceedings{moon2018v2v,
  title={V2v-posenet: Voxel-to-voxel prediction network for accurate 3d hand and human pose estimation from a single depth map},
  author={Moon, Gyeongsik and Chang, Ju Yong and Lee, Kyoung Mu},
  booktitle={Proceedings of the IEEE/CVF Conference on Computer Vision and Pattern Recognition},
  pages={5079--5088},
  year={2018}
}

@inproceedings{ge2018point,
  title={Point-to-point regression pointnet for 3d hand pose estimation},
  author={Ge, Liuhao and Ren, Zhou and Yuan, Junsong},
  booktitle={Proceedings of the European Conference on Computer Vision},
  pages={475--491},
  year={2018}
}

@inproceedings{cheng2021handfoldingnet,
  title={Handfoldingnet: A 3d hand pose estimation network using multiscale-feature guided folding of a 2d hand skeleton},
  author={Cheng, Wencan and Park, Jae Hyun and Ko, Jong Hwan},
  booktitle={Proceedings of the IEEE/CVF International Conference on Computer Vision},
  pages={11260--11269},
  year={2021}
}

@inproceedings{li2023renderih,
  title={RenderIH: A large-scale synthetic dataset for 3d interacting hand pose estimation},
  author={Li, Lijun and Tian, Linrui and Zhang, Xindi and Wang, Qi and Zhang, Bang and Bo, Liefeng and Liu, Mengyuan and Chen, Chen},
  booktitle={Proceedings of the IEEE/CVF International Conference on Computer Vision},
  pages={20395--20405},
  year={2023}
}

@inproceedings{gong2023diffpose,
  title={Diffpose: Toward more reliable 3d pose estimation},
  author={Gong, Jia and Foo, Lin Geng and Fan, Zhipeng and Ke, Qiuhong and Rahmani, Hossein and Liu, Jun},
  booktitle={Proceedings of the IEEE/CVF Conference on Computer Vision and Pattern Recognition},
  pages={13041--13051},
  year={2023}
}

@inproceedings{ge2018hand,
  title={Hand pointnet: 3d hand pose estimation using point sets},
  author={Ge, Liuhao and Cai, Yujun and Weng, Junwu and Yuan, Junsong},
  booktitle={Proceedings of the IEEE/CVF Conference on Computer Vision and Pattern Recognition},
  pages={8417--8426},
  year={2018}
}

@inproceedings{hampali2020honnotate,
  title={Honnotate: A method for 3d annotation of hand and object poses},
  author={Hampali, Shreyas and Rad, Mahdi and Oberweger, Markus and Lepetit, Vincent},
  booktitle={Proceedings of the IEEE/CVF Conference on Computer Vision and Pattern Recognition},
  pages={3196--3206},
  year={2020}
}

@article{tompson2014real,
  title={Real-time continuous pose recovery of human hands using convolutional networks},
  author={Tompson, Jonathan and Stein, Murphy and Lecun, Yann and Perlin, Ken},
  journal={ACM Transactions on Graphics (ToG)},
  volume={33},
  number={5},
  pages={1--10},
  year={2014},
  publisher={ACM New York, NY, USA}
}

@article{zhu2020deformable,
  title={Deformable detr: Deformable transformers for end-to-end object detection},
  author={Zhu, Xizhou and Su, Weijie and Lu, Lewei and Li, Bin and Wang, Xiaogang and Dai, Jifeng},
  journal={arXiv preprint arXiv:2010.04159},
  year={2020}
}

@article{song2020denoising,
  title={Denoising diffusion implicit models},
  author={Song, Jiaming and Meng, Chenlin and Ermon, Stefano},
  journal={arXiv preprint arXiv:2010.02502},
  year={2020}
}

@inproceedings{yang2022artiboost,
  title={ArtiBoost: Boosting articulated 3d hand-object pose estimation via online exploration and synthesis},
  author={Yang, Lixin and Li, Kailin and Zhan, Xinyu and Lv, Jun and Xu, Wenqiang and Li, Jiefeng and Lu, Cewu},
  booktitle={Proceedings of the IEEE/CVF Conference on Computer Vision and Pattern Recognition},
  pages={2750--2760},
  year={2022}
}

@inproceedings{park2022handoccnet,
  title={Handoccnet: Occlusion-robust 3d hand mesh estimation network},
  author={Park, JoonKyu and Oh, Yeonguk and Moon, Gyeongsik and Choi, Hongsuk and Lee, Kyoung Mu},
  booktitle={Proceedings of the IEEE/CVF Conference on Computer Vision and Pattern Recognition},
  pages={1496--1505},
  year={2022}
}

@inproceedings{moon2020i2l,
  title={I2l-meshnet: Image-to-lixel prediction network for accurate 3d human pose and mesh estimation from a single rgb image},
  author={Moon, Gyeongsik and Lee, Kyoung Mu},
  booktitle={Proceedings of the European Conference on Computer Vision},
  pages={752--768},
  year={2020},
  organization={Springer}
}

@inproceedings{ho2020denoising,
  title={Denoising diffusion probabilistic models},
  author={Ho, Jonathan and Jain, Ajay and Abbeel, Pieter},
  booktitle={Proceedings of the Advances in Neural Information Processing Systems},
  volume={33},
  pages={6840--6851},
  year={2020}
}

@inproceedings{nichol2021improved,
  title={Improved denoising diffusion probabilistic models},
  author={Nichol, Alexander Quinn and Dhariwal, Prafulla},
  booktitle={Proceedings of the International Conference on Machine Learning},
  pages={8162--8171},
  year={2021},
  $organization={PMLR}
}

@inproceedings{tang2014latent,
  title={Latent regression forest: Structured estimation of 3d articulated hand posture},
  author={Tang, Danhang and Jin Chang, Hyung and Tejani, Alykhan and Kim, Tae-Kyun},
  booktitle={Proceedings of the IEEE/CVF Conference on Computer Vision and Pattern Recognition},
  pages={3786--3793},
  year={2014}
}

@article{romero2022embodied,
  title={Embodied hands: Modeling and capturing hands and bodies together},
  author={Romero, Javier and Tzionas, Dimitrios and Black, Michael J},
  journal={arXiv preprint arXiv:2201.02610},
  year={2022}
}

@inproceedings{ge20193d,
  title={3d hand shape and pose estimation from a single rgb image},
  author={Ge, Liuhao and Ren, Zhou and Li, Yuncheng and Xue, Zehao and Wang, Yingying and Cai, Jianfei and Yuan, Junsong},
  booktitle={Proceedings of the IEEE/CVF Conference on Computer Vision and Pattern Recognition},
  pages={10833--10842},
  year={2019}
}

@inproceedings{wang2023memahand,
  title={MeMaHand: Exploiting mesh-mano interaction for single image two-hand reconstruction},
  author={Wang, Congyi and Zhu, Feida and Wen, Shilei},
  booktitle={Proceedings of the IEEE/CVF Conference on Computer Vision and Pattern Recognition},
  pages={564--573},
  year={2023}
}

@article{deng2022recurrent,
  title={Recurrent 3D hand pose estimation using cascaded pose-guided 3D alignments},
  author={Deng, Xiaoming and Zuo, Dexin and Zhang, Yinda and Cui, Zhaopeng and Cheng, Jian and Tan, Ping and Chang, Liang and Pollefeys, Marc and Fanello, Sean and Wang, Hongan},
  journal={IEEE Transactions on Pattern Analysis and Machine Intelligence},
  volume={45},
  number={1},
  pages={932--945},
  year={2022},
  publisher={IEEE}
}

@inproceedings{he2016deep,
  title={Deep residual learning for image recognition},
  author={He, Kaiming and Zhang, Xiangyu and Ren, Shaoqing and Sun, Jian},
  booktitle={Proceedings of the IEEE/CVF Conference on Computer Vision and Pattern Recognition},
  pages={770--778},
  year={2016}
}

@inproceedings{ren2023decoupled,
  title={Decoupled iterative refinement framework for interacting hands reconstruction from a single rgb image},
  author={Ren, Pengfei and Wen, Chao and Zheng, Xiaozheng and Xue, Zhou and Sun, Haifeng and Qi, Qi and Wang, Jingyu and Liao, Jianxin},
  booktitle={Proceedings of the IEEE/CVF International Conference on Computer Vision},
  pages={8014--8025},
  year={2023}
}

@inproceedings{wang2023interacting,
  title={Interacting hand-object pose estimation via dense mutual attention},
  author={Wang, Rong and Mao, Wei and Li, Hongdong},
  booktitle={Proceedings of the IEEE/CVF Winter Conference on Applications of Computer Vision},
  pages={5735--5745},
  year={2023}
}

@inproceedings{qi2024hoisdf,
  title={{HOISDF}: Constraining {3D} Hand-Object Pose Estimation with Global Signed Distance Fields},
  author={Qi, Haozhe and Zhao, Chen and Salzmann, Mathieu and Mathis, Alexander},
  booktitle={Proceedings of the IEEE/CVF Conference on Computer Vision and Pattern Recognition (CVPR)},
  month={June},
  pages={10392--10402},
  year={2024}
}

@InProceedings{huang2023nvf,
    author    = {Huang, Lin and Lin, Chung-Ching and Lin, Kevin and Liang, Lin and Wang, Lijuan and Yuan, Junsong and Liu, Zicheng},
    title     = {Neural voting field for camera-space 3d hand pose estimation},
    booktitle = {Proceedings of the IEEE/CVF Conference on Computer Vision and Pattern Recognition},
    month     = {June},
    year      = {2023},
    pages     = {8969-8978}
}

@inproceedings{du2019crossinfonet,
  title={Crossinfonet: Multi-task information sharing based hand pose estimation},
  author={Du, Kuo and Lin, Xiangbo and Sun, Yi and Ma, Xiaohong},
  booktitle={Proceedings of the IEEE/CVF Conference on Computer Vision and Pattern Recognition},
  pages={9896--9905},
  year={2019}
}

@inproceedings{cheng2022efficient,
  title={Efficient virtual view selection for 3d hand pose estimation},
  author={Cheng, Jian and Wan, Yanguang and Zuo, Dexin and Ma, Cuixia and Gu, Jian and Tan, Ping and Wang, Hongan and Deng, Xiaoming and Zhang, Yinda},
  booktitle={Proceedings of the AAAI Conference on Artificial Intelligence},
  volume={36},
  number={1},
  pages={419--426},
  year={2022}
}

@inproceedings{huang2020awr,
  title={Awr: Adaptive weighting regression for 3d hand pose estimation},
  author={Huang, Weiting and Ren, Pengfei and Wang, Jingyu and Qi, Qi and Sun, Haifeng},
  booktitle={Proceedings of the AAAI Conference on Artificial Intelligence},
  volume={34},
  number={07},
  pages={11061--11068},
  year={2020}
}

@article{wang2018mask,
  title={Mask-pose cascaded cnn for 2d hand pose estimation from single color image},
  author={Wang, Yangang and Peng, Cong and Liu, Yebin},
  journal={IEEE Transactions on Circuits and Systems for Video Technology},
  volume={29},
  number={11},
  pages={3258--3268},
  year={2018},
  publisher={IEEE}
}

@article{jiang20253d,
  title={{3D} Hand Pose Estimation via Articulated Anchor-to-Joint {3D} Local Regressors},
  author={Jiang, Changlong and Xiao, Yang and Zheng, Jinghong and Kuang, Haohong and Wu, Cunlin and Zhang, Mingyang and Cao, Zhiguo and Du, Min and Zhou, Joey Tianyi and Yuan, Junsong},
  journal={IEEE Transactions on Pattern Analysis and Machine Intelligence},
  volume={48},
  number={1},
  pages={982--998},
  year={2026},
  doi={10.1109/TPAMI.2025.3609907}
}

@inproceedings{cheng2024handdiff,
  title={Handdiff: 3d hand pose estimation with diffusion on image-point cloud},
  author={Cheng, Wencan and Tang, Hao and Van Gool, Luc and Ko, Jong Hwan},
  booktitle={Proceedings of the IEEE/CVF Conference on Computer Vision and Pattern Recognition},
  pages={2274--2284},
  year={2024}
}

@inproceedings{zhou2026df,
  title={DF-Mamba: Deformable State Space Modeling for 3D Hand Pose Estimation in Interactions},
  author={Zhou, Yifan and Ohkawa, Takehiko and Zhou, Guwenxiao and Goto, Kanoko and Hirose, Takumi and Sekikawa, Yusuke and Inoue, Nakamasa},
  booktitle={2026 IEEE/CVF Winter Conference on Applications of Computer Vision (WACV)},
  pages={5352--5363},
  year={2026},
  organization={IEEE}
}

@article{zheng2026pandapose,
  title={PandaPose: 3D Human Pose Lifting from a Single Image via Propagating 2D Pose Prior to 3D Anchor Space},
  author={Zheng, Jinghong and Jiang, Changlong and Xiao, Yang and Li, Jiaqi and Kuang, Haohong and Xu, Hang and Wang, Ran and Cao, Zhiguo and Du, Min and Zhou, Joey Tianyi},
  journal={Advances in Neural Information Processing Systems},
  volume={38},
  pages={90147--90169},
  year={2026}
}

@inproceedings{kuang2026jump,
  title={JUMP-Hand: Learning Joint-wise Uncertainty to Gate Mixture of View Experts for Multi-View 3D Hand Reconstruction},
  author={Kuang, Haohong and Xiao, Yang and Jiang, Changlong and Zheng, Jinghong and Xu, Hang and Wang, Ran and Cao, Zhiguo and Zhou, Joey Tianyi},
  booktitle={Proceedings of the IEEE/CVF Conference on Computer Vision and Pattern Recognition},
  pages={28348--28357},
  year={2026}
}

@inproceedings{xu2026moeg,
  title={MoEG-HOI: Mixture of Expert Groups for One-Stage Hand-Object Interaction Motion Generation with Hand-Finger-Joint Semantic Guidance},
  author={Xu, Hang and Xiao, Yang and Jiang, Changlong and Kuang, Haohong and Zhang, Kaidi and Du, Min and Wang, Ran},
  booktitle={Proceedings of the AAAI Conference on Artificial Intelligence},
  volume={40},
  number={13},
  pages={11223--11231},
  year={2026}
}

@article{tan2024epmf,
  title={EPMF: Efficient perception-aware multi-sensor fusion for 3D semantic segmentation},
  author={Tan, Mingkui and Zhuang, Zhuangwei and Chen, Sitao and Li, Rong and Jia, Kui and Wang, Qicheng and Li, Yuanqing},
  journal={IEEE Transactions on Pattern Analysis and Machine Intelligence},
  volume={46},
  number={12},
  pages={8258--8273},
  year={2024},
  publisher={IEEE}
}

@article{ding2023image,
  title={Image captioning with controllable and adaptive length levels},
  author={Ding, Ning and Deng, Chaorui and Tan, Mingkui and Du, Qing and Ge, Zhiwei and Wu, Qi},
  journal={IEEE Transactions on Pattern Analysis and Machine Intelligence},
  volume={46},
  number={2},
  pages={764--779},
  year={2023},
  publisher={IEEE}
}

@inproceedings{huang2020location,
  title={Location-aware graph convolutional networks for video question answering},
  author={Huang, Deng and Chen, Peihao and Zeng, Runhao and Du, Qing and Tan, Mingkui and Gan, Chuang},
  booktitle={Proceedings of the AAAI conference on artificial intelligence},
  volume={34},
  number={07},
  pages={11021--11028},
  year={2020}
}

@inproceedings{niu2022efficient,
  title={Efficient test-time model adaptation without forgetting},
  author={Niu, Shuaicheng and Wu, Jiaxiang and Zhang, Yifan and Chen, Yaofo and Zheng, Shijian and Zhao, Peilin and Tan, Mingkui},
  booktitle={International conference on machine learning},
  pages={16888--16905},
  year={2022},
  organization={PMLR}
}

@article{zhang2020collaborative,
  title={Collaborative unsupervised domain adaptation for medical image diagnosis},
  author={Zhang, Yifan and Wei, Ying and Wu, Qingyao and Zhao, Peilin and Niu, Shuaicheng and Huang, Junzhou and Tan, Mingkui},
  journal={IEEE Transactions on Image Processing},
  volume={29},
  pages={7834--7844},
  year={2020},
  publisher={IEEE}
}

@article{liu2021discrimination,
  title={Discrimination-aware network pruning for deep model compression},
  author={Liu, Jing and Zhuang, Bohan and Zhuang, Zhuangwei and Guo, Yong and Huang, Junzhou and Zhu, Jinhui and Tan, Mingkui},
  journal={IEEE Transactions on Pattern Analysis and Machine Intelligence},
  volume={44},
  number={8},
  pages={4035--4051},
  year={2021},
  publisher={IEEE}
}

\end{document}